\documentclass[letter, 10pt, conference]{ieeeconf}      

\IEEEoverridecommandlockouts                              

\usepackage{graphics} 
\usepackage{epsfig} 
\usepackage{mathptmx} 
\usepackage{times} 
\usepackage{amsmath} 
\usepackage{amssymb}  
\usepackage{xcolor}
\usepackage{pythonhighlight}
\usepackage{caption} 
\usepackage{subcaption}
\usepackage{graphicx} 
\usepackage{pifont} 
\usepackage{bmpsize}
\usepackage{xspace}
\usepackage[utf8]{inputenc} 
\usepackage[T1]{fontenc}    
\usepackage{booktabs}
\usepackage{url} 
\usepackage{adjustbox}
\usepackage[hidelinks]{hyperref} 
\hypersetup{
    colorlinks,
    linkcolor={blue!50!black},
    citecolor={blue!50!black},
    urlcolor={blue!80!black}
} 
\usepackage{cleveref}
\usepackage{amsfonts}       
\usepackage{nicefrac}       
\usepackage{stfloats}
\usepackage{microtype}      
\usepackage{float}
\usepackage{placeins}
\usepackage{multirow}
\usepackage{bbold}
\usepackage{algorithm}
\usepackage{algpseudocode}
\usepackage{xspace}

\DeclareTextFontCommand{\emph}{\em}
\newcommand{\waymax}{Waymax\xspace}

\newcommand{\cmark}{\ding{51}}%

\newcommand{\tde}{$\texttt{TorchDriveEnv}$\xspace}
\newcommand{\tds}{$\texttt{TorchDriveSim}$\xspace}  

\newcommand{\initlink}{\href{https://docs.inverted.ai/en/latest/pythonapi/sdk-initialize.html}{INITIALIZE}\xspace} 

\newcommand{\drivelink}{\href{https://docs.inverted.ai/en/latest/}{DRIVE}\xspace}

\title{\LARGE \bf
TorchDriveEnv: A Reinforcement Learning Benchmark for Autonomous Driving with Reactive, Realistic, and Diverse Non-Playable Characters
}

\author{
Jonathan Wilder Lavington$^{1,2,*}$ %
Ke Zhang$^{1,2,*}$ %
Vasileios Lioutas$^{1,2}$ 
Matthew Niedoba$^{1,2}$%
\\ %
Yunpeng Liu$^{1,2}$ %
Dylan Green$^{1,2}$ %
Saeid Naderiparizi$^{1,2}$ %
Xiaoxuan Liang$^{1,2}$ 
Setareh Dabiri$^{1}$ %
\\ %
Adam \'Scibior$^{1,2}$ %
Berend Zwartsenberg$^{1}$ %
Frank Wood$^{1,2,3}$ %
\thanks{\!\!\!\!\!\!$^{1}$Inverted AI, $^{2}$University of British Columbia, $^{3}$MILA \newline
$^{*}$Denotes equal contribution}%
}

\begin{document}

\maketitle
\thispagestyle{empty}
\pagestyle{empty}

 
\begin{abstract} 
The training, testing, and deployment, of autonomous vehicles requires realistic and efficient simulators. Moreover, because of the high variability between different problems presented in different autonomous systems, these simulators need to be easy to use, and easy to modify. To address these problems we introduce \tds and its benchmark extension \tde. \tde is a lightweight reinforcement learning benchmark programmed entirely in Python, which can be modified to test a number of different factors in learned vehicle behaviour, including the effect of varying kinematic models, agent types, and traffic control patterns. Most importantly unlike many replay based simulation approaches, \tde is fully integrated with a state of the art behavioural simulation API. This allows users to train and evaluate driving models alongside data driven Non-Playable Characters (NPC) whose initializations and driving behaviour are reactive, realistic, and diverse. We illustrate the efficiency and simplicity of \tde by evaluating common reinforcement learning baselines in both training and validation environments. Our experiments show that \tde is easy to use, but difficult to solve.
\end{abstract}


\begin{figure}[ht!]
    \centering 
    \begin{subfigure}{0.4865\textwidth}
       \includegraphics[width=0.19\textwidth]{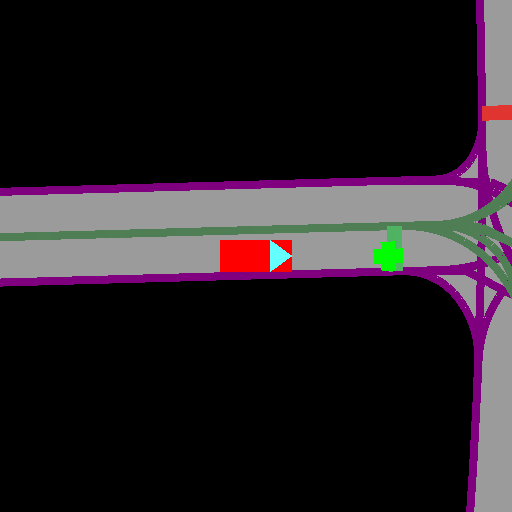}
       \includegraphics[width=0.19\textwidth]{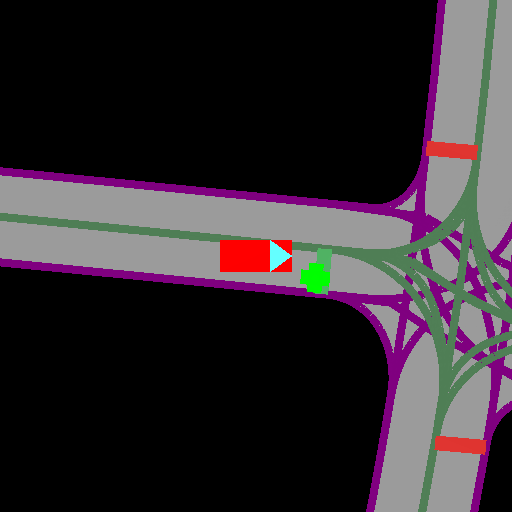} 
       \includegraphics[width=0.19\textwidth]{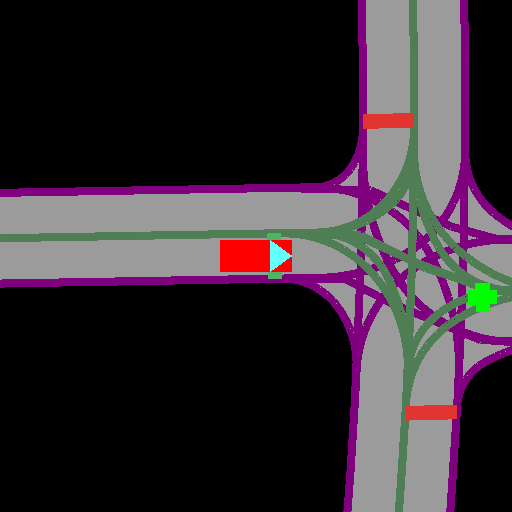} 
       \includegraphics[width=0.19\textwidth]{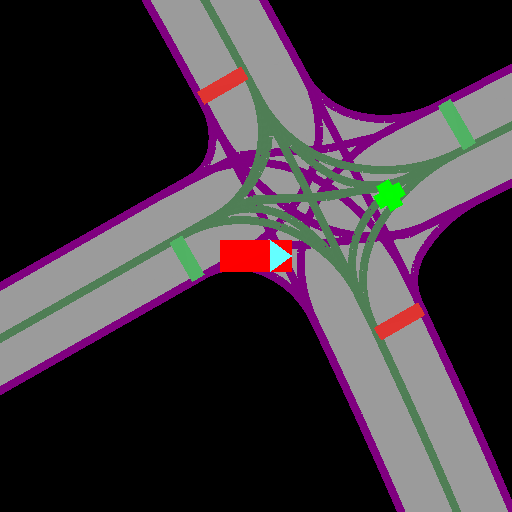} 
       \includegraphics[width=0.19\textwidth]{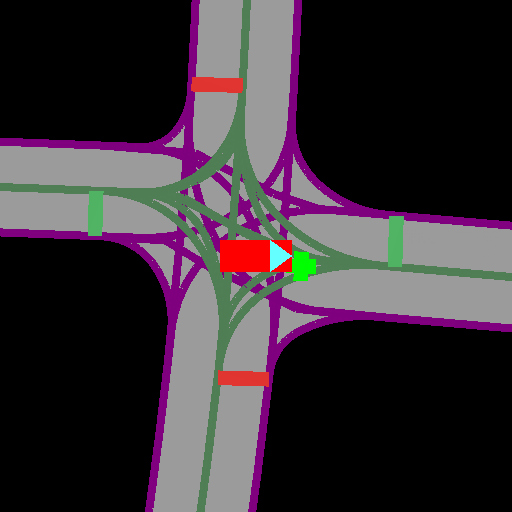}  
       \caption{Empty Intersection } \vspace{0.1cm}
       \label{fig:train-ex:empty-i} 
    \end{subfigure} 
    \begin{subfigure}{0.4865\textwidth}
       \includegraphics[width=0.19\textwidth]{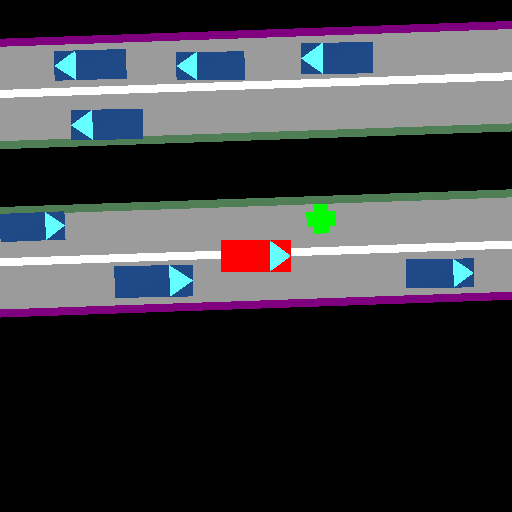}
       \includegraphics[width=0.19\textwidth]{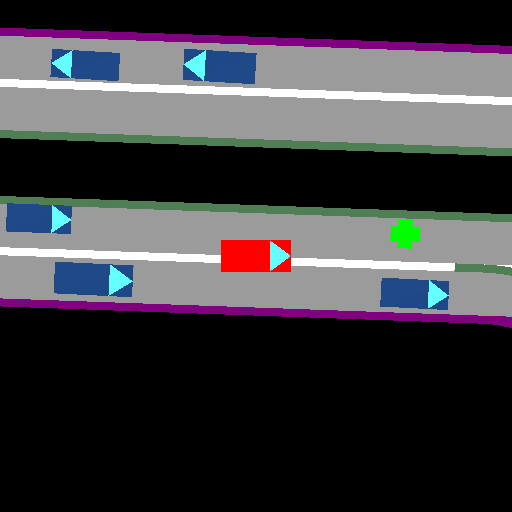} 
       \includegraphics[width=0.19\textwidth]{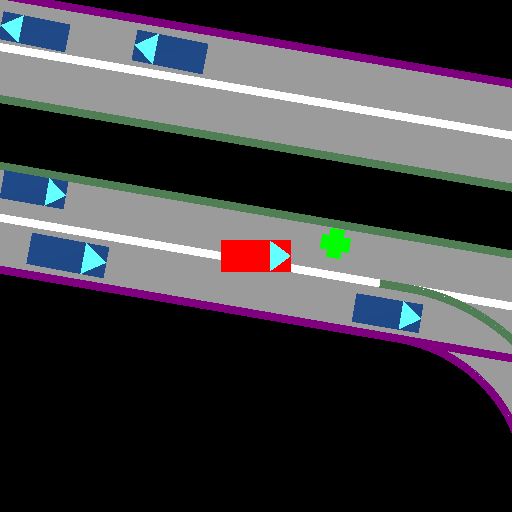} 
       \includegraphics[width=0.19\textwidth]{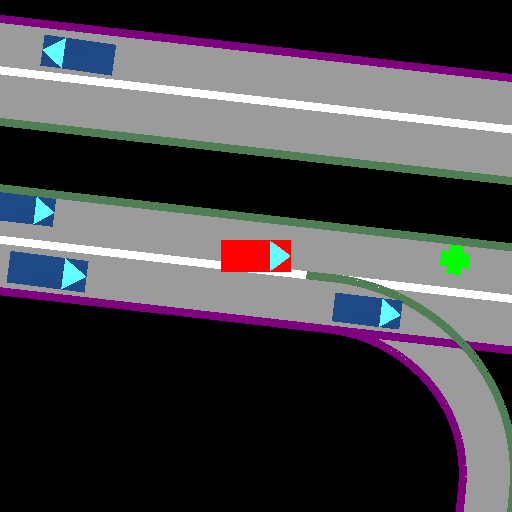} 
       \includegraphics[width=0.19\textwidth]{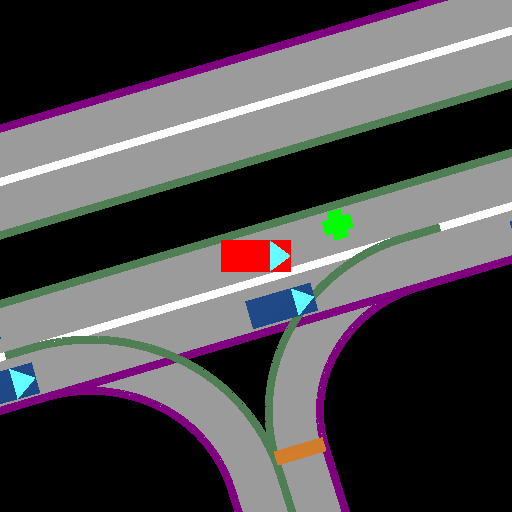}  
       \caption{Crowded Highway } \vspace{0.1cm}
       \label{fig:train-ex:crowded-h}  
    \end{subfigure} 
    \begin{subfigure}{0.4865\textwidth}
      \includegraphics[width=0.19\textwidth]{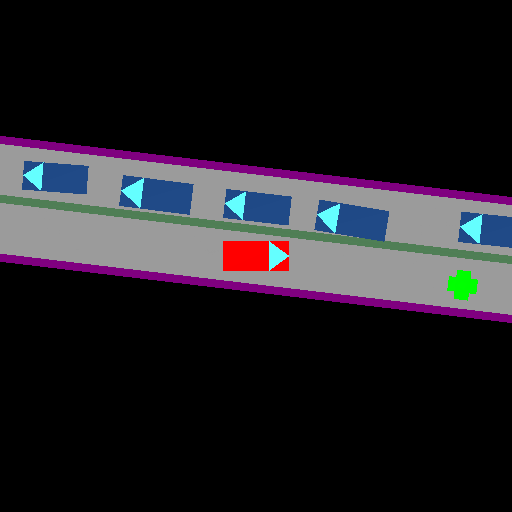}
       \includegraphics[width=0.19\textwidth]{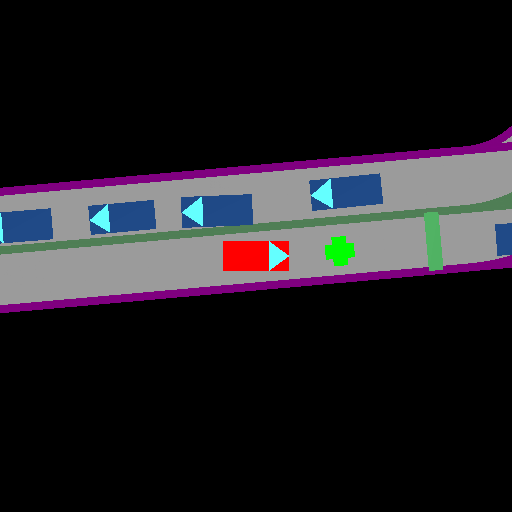} 
       \includegraphics[width=0.19\textwidth]{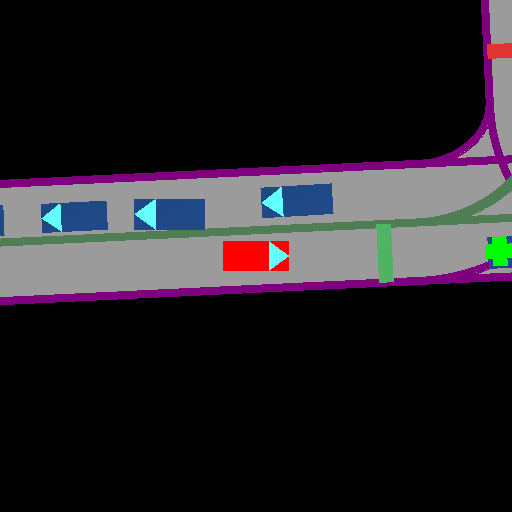} 
       \includegraphics[width=0.19\textwidth]{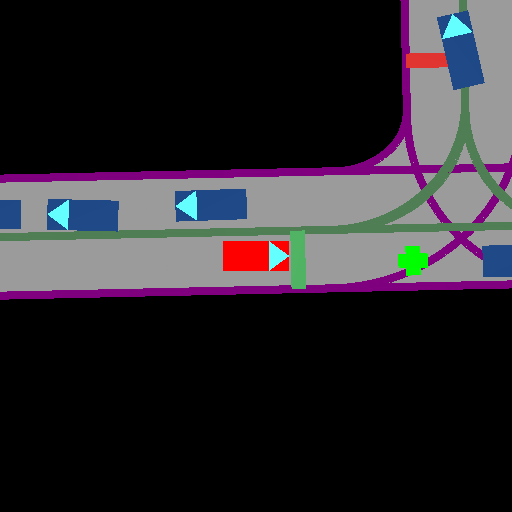} 
       \includegraphics[width=0.19\textwidth]{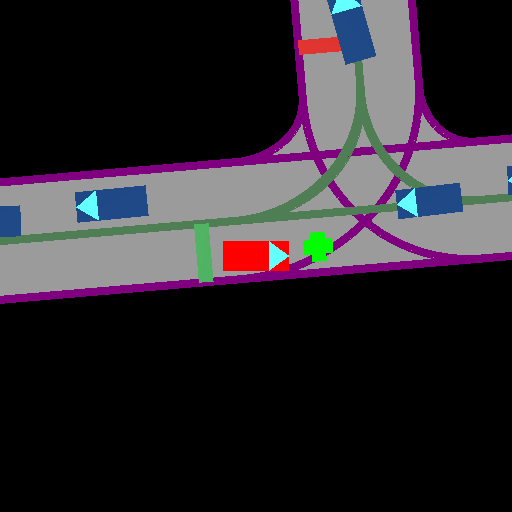} 
       \caption{Crowded Left Turn } \vspace{0.1cm}
       \label{fig:train-ex:crowded-i}
    \end{subfigure}
    \begin{subfigure}{0.4865\textwidth}
       \includegraphics[width=0.19\textwidth]{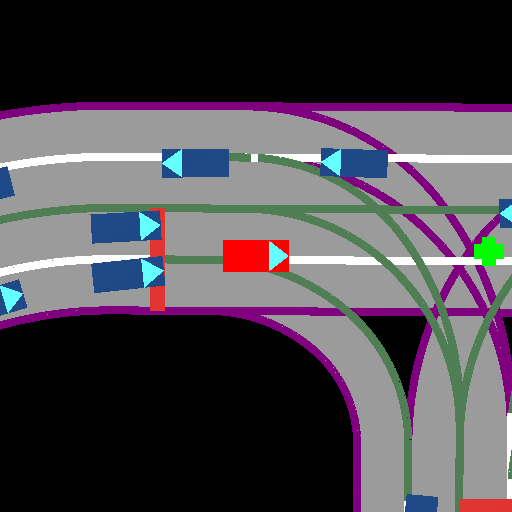}
       \includegraphics[width=0.19\textwidth]{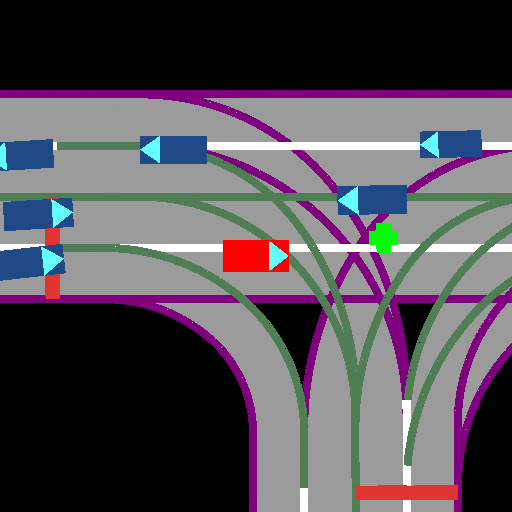} 
       \includegraphics[width=0.19\textwidth]{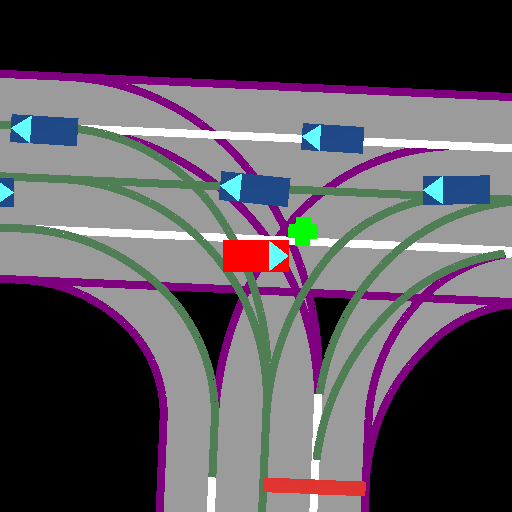} 
       \includegraphics[width=0.19\textwidth]{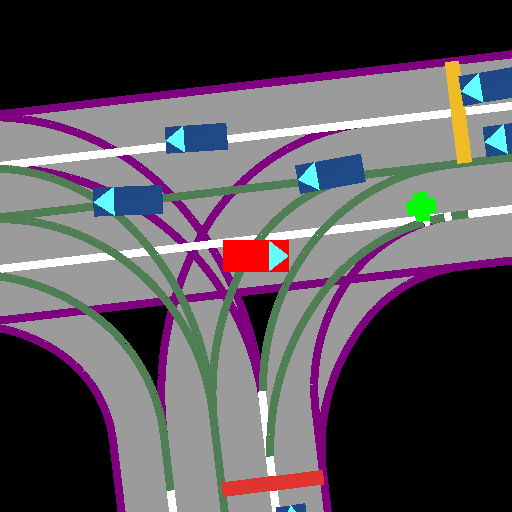} 
       \includegraphics[width=0.19\textwidth]{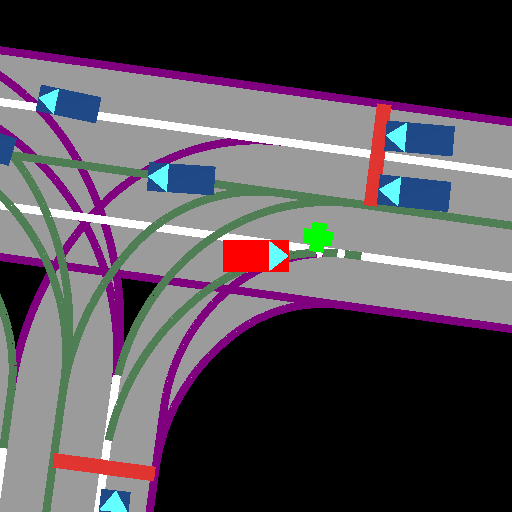}  
       \caption{Controlled Intersection } \vspace{0.1cm} 
       \label{fig:train-ex:crowded-m}
    \end{subfigure}
    \begin{subfigure}{0.4865\textwidth} 
       \includegraphics[width=0.19\textwidth]{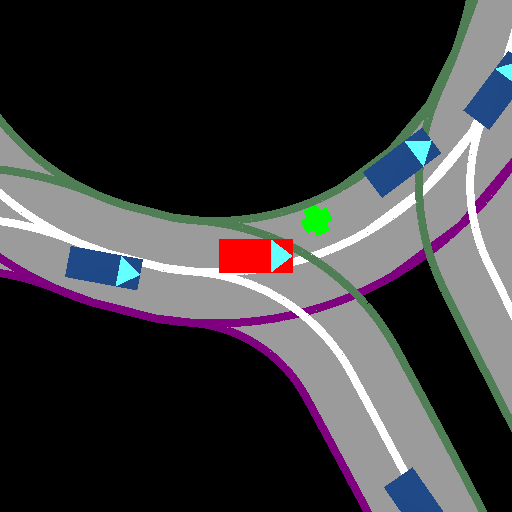}
       \includegraphics[width=0.19\textwidth]{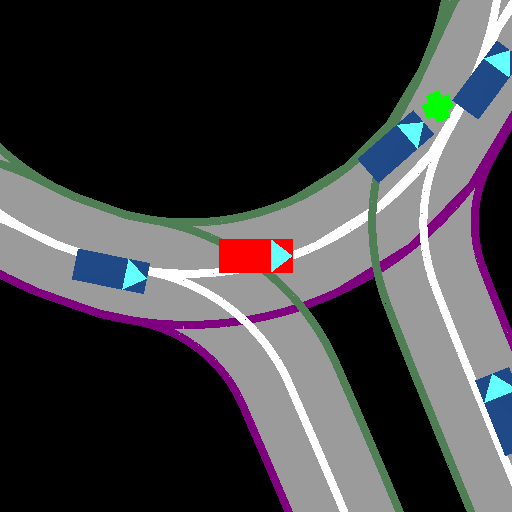} 
       \includegraphics[width=0.19\textwidth]{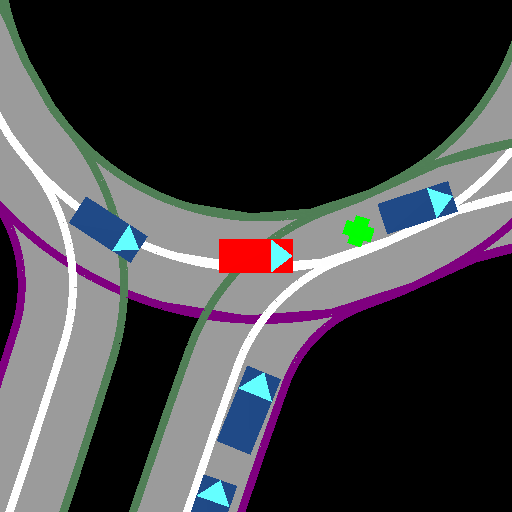} 
       \includegraphics[width=0.19\textwidth]{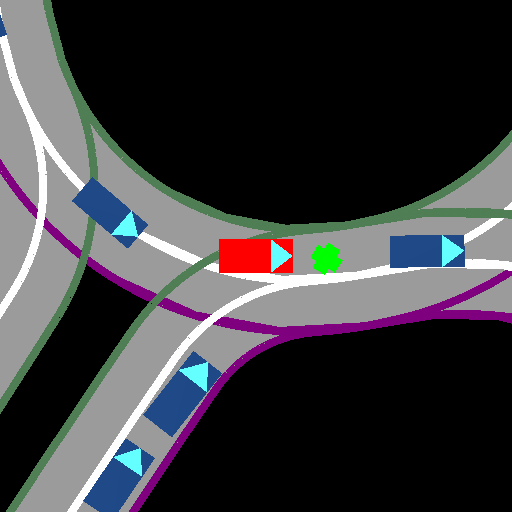} 
       \includegraphics[width=0.19\textwidth]{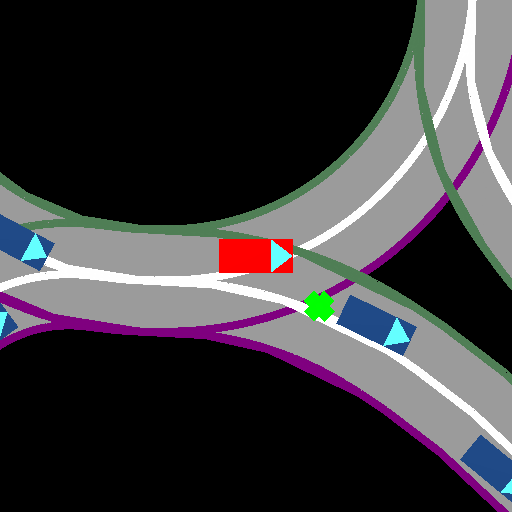}  
       \caption{Roundabout }  
       \label{fig:train-ex:crowded-r}
    \end{subfigure}
    \caption{Frames from five  \tde  stochastic episodes encountered during  SAC~\cite{haarnoja2018soft} RL agent training. In all examples, the ego vehicle is red, the next waypoint is a green circle, and non-playable character (NPC) vehicles  are  blue. Drivable surfaces are grey, while traffic lights are denoted by thin coloured rectangles at stop-lines that are red, yellow, or green. Lanes are indicated by purple, dark green and white lines for right, left, or overlapping lane boundaries. In each example, five uniformly spaced frames were taken from a single episode and are displayed sequentially left to right. These examples illustrate the diverse traffic and road conditions encountered during training. Note the high density and realistic behaviour displayed by the NPCs, particularly in \Cref{fig:train-ex:crowded-h,fig:train-ex:crowded-i,fig:train-ex:crowded-r,fig:train-ex:crowded-m}. 
    } 
    \vspace{-0.9cm} 
    \label{fig:train-ex}
\end{figure}

\begin{figure*}[ht]
    \centering
    \begin{subfigure}{0.19\textwidth}
       \includegraphics[width=\textwidth, height=\textwidth]{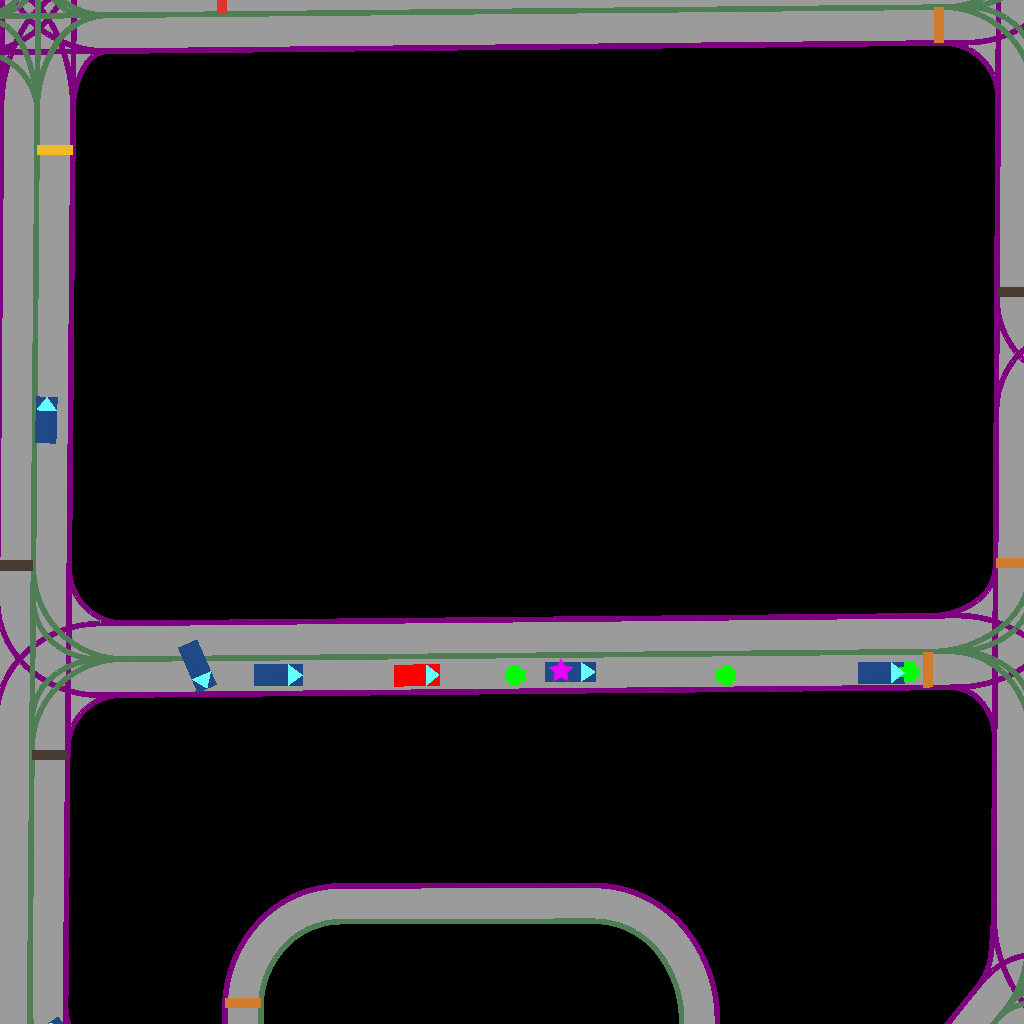}
       \caption{Parked-Car}
       \label{fig:val-ex:1}
    \end{subfigure}  \hfill
    \begin{subfigure}{0.19\textwidth}
       \includegraphics[width=\textwidth, height=\textwidth]{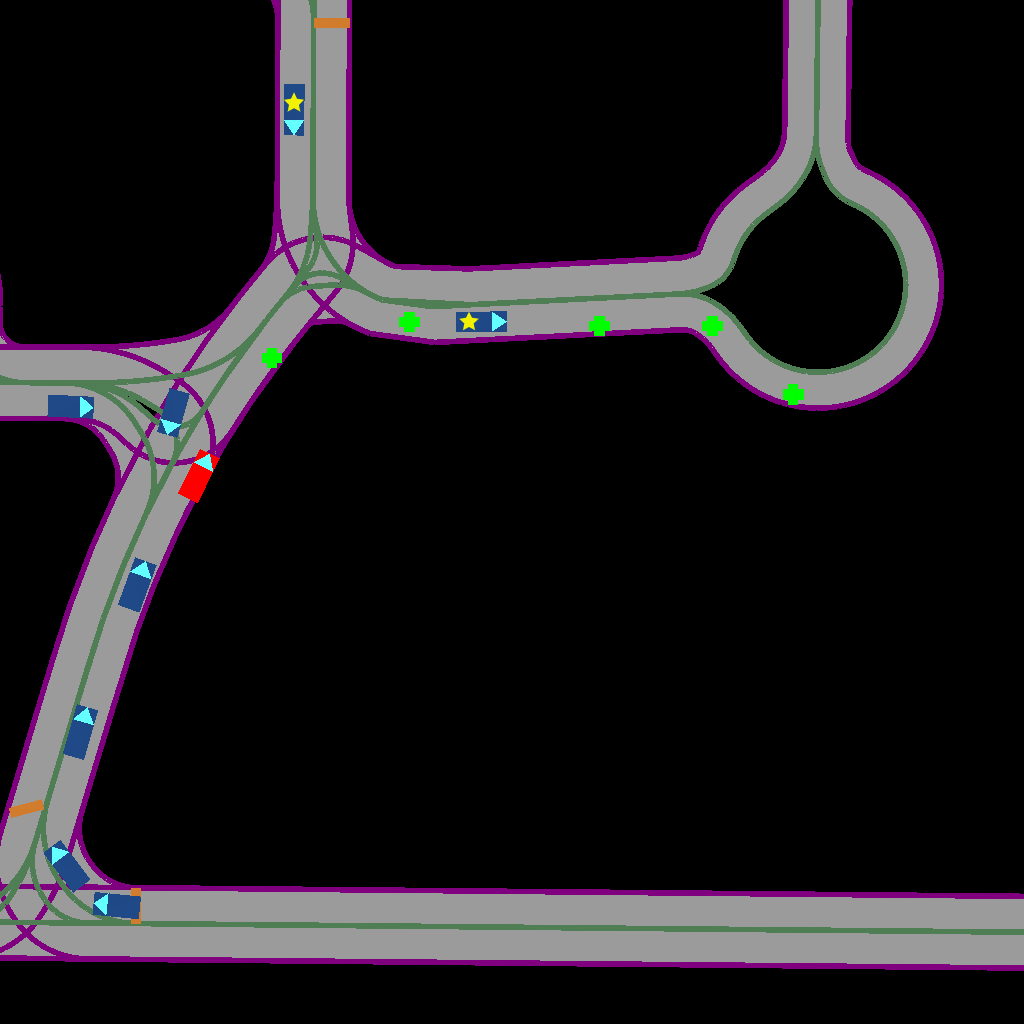} 
       \caption{Three-Way}
       \label{fig:val-ex:2}
    \end{subfigure} \hfill
    \begin{subfigure}{0.19\textwidth}
       \includegraphics[width=\textwidth, height=\textwidth]{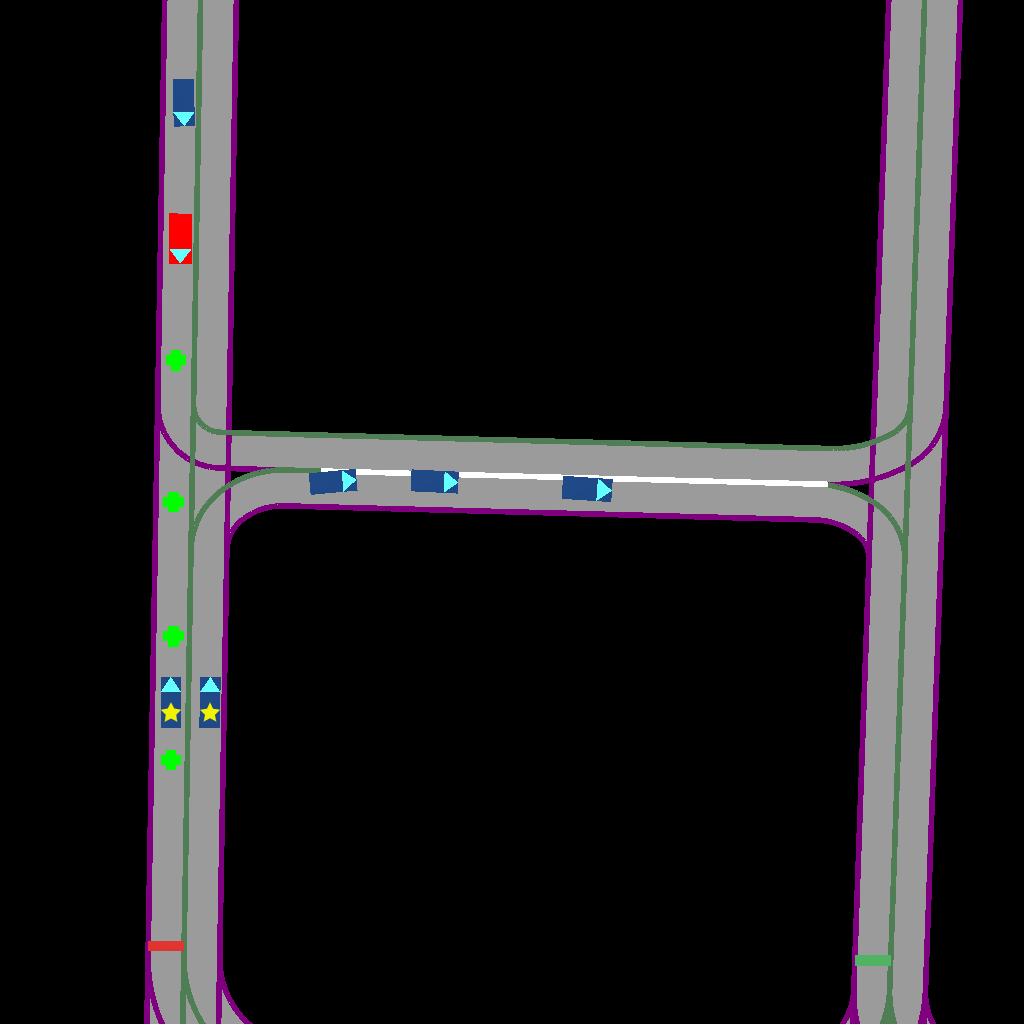} 
       \caption{Chicken}
       \label{fig:val-ex:3}
    \end{subfigure}  \hfill
    \begin{subfigure}{0.19\textwidth}
       \includegraphics[width=\textwidth, height=\textwidth]{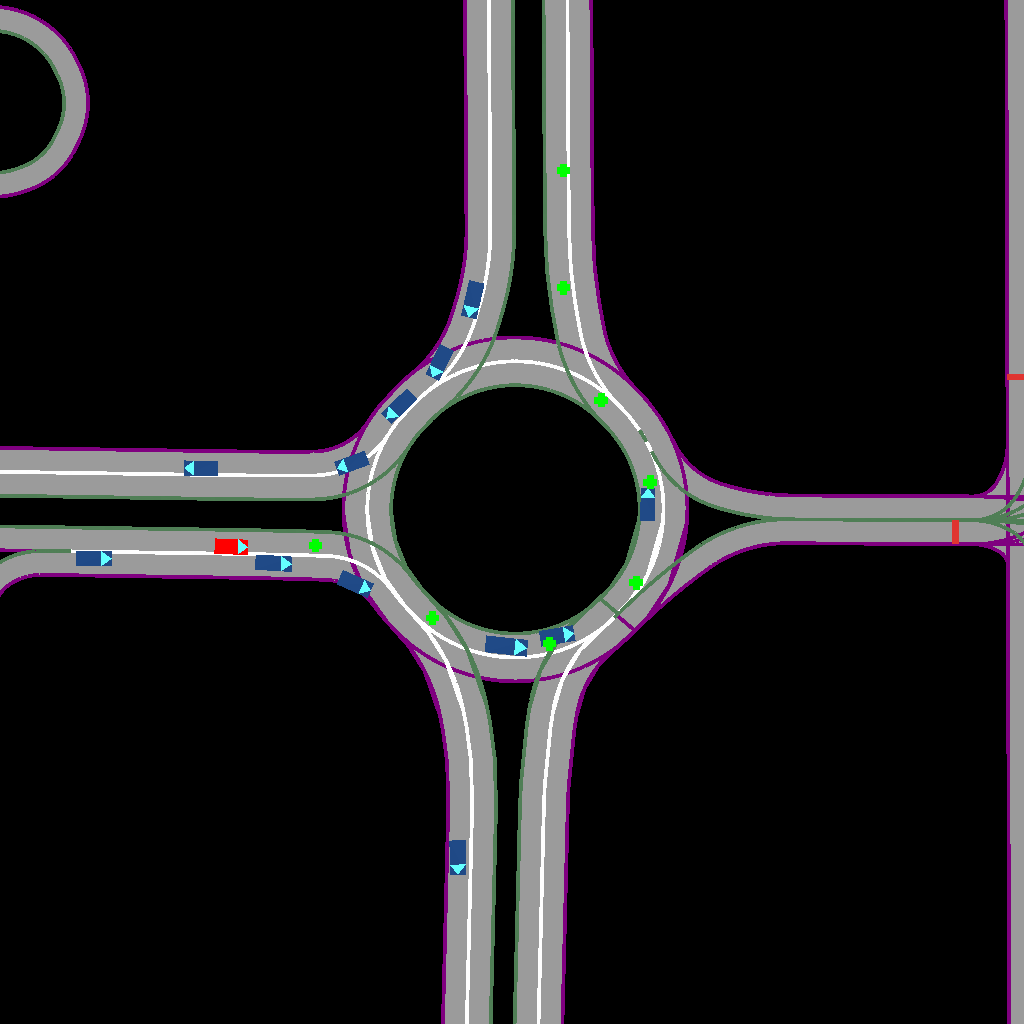} 
       \caption{Roundabout}
       \label{fig:val-ex:4}
    \end{subfigure} \hfill
    \begin{subfigure}{0.19\textwidth}
       \includegraphics[width=\textwidth, height=\textwidth]{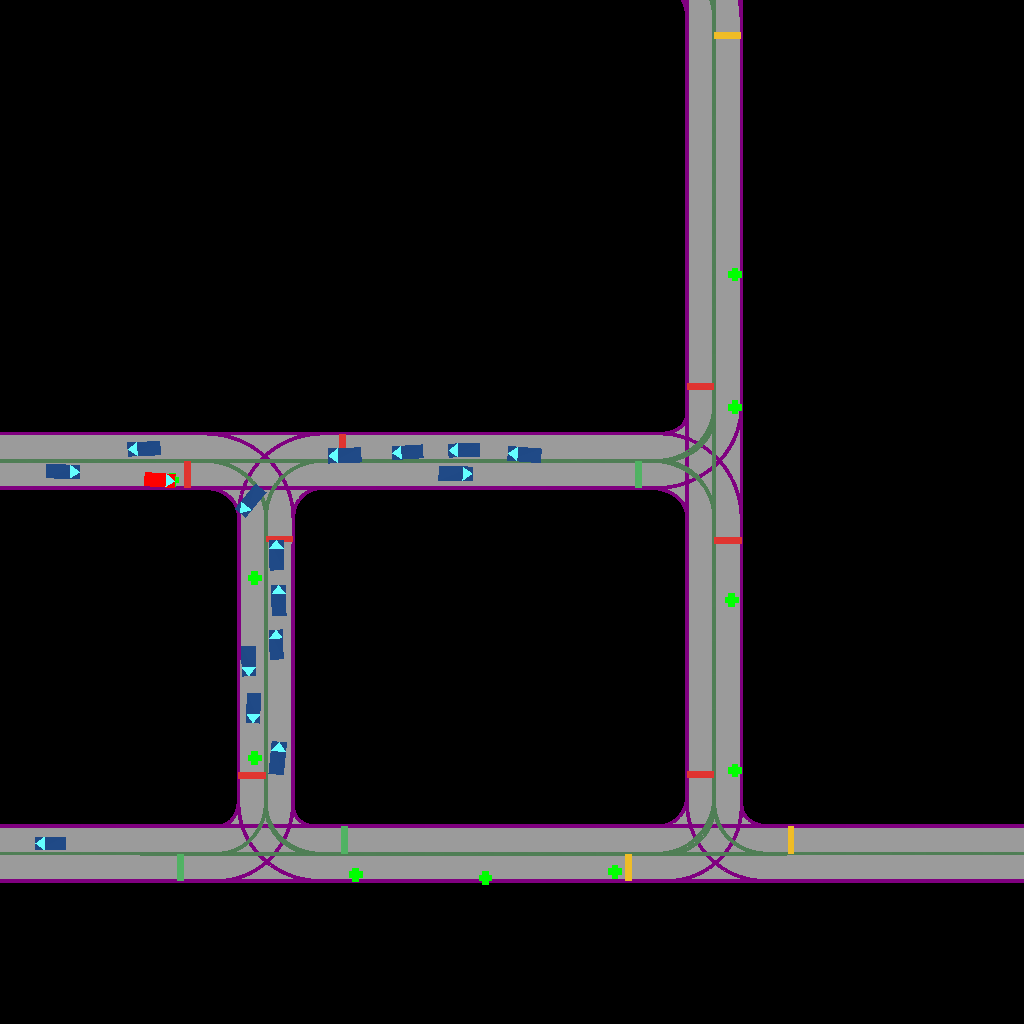} 
       \caption{Traffic-Lights}
       \label{fig:val-ex:5}
    \end{subfigure}   
    \caption{Stochastic initializations of validation scenarios used to test learned agents' out of distribution performance. Each scenario tests a different set of agent capabilities.  The first scenario (a) tests an agent's ability to drive around a parked vehicle obstructing the road. The three-way intersection (b) tests an agent's ability to navigate a three way intersection and yield as needed to cross traffic. The example in (c) requires the agent to negotiate around a collision with an oncoming passing car in its lane. The roundabout in (d) requires the agent to merge, yield, change lanes, and finally, exit the roundabout. The last example in (e) tests if the agent can navigate through controlled intersections in the presence of NPCs that obey traffic lights realistically.}
    \label{fig:val-ex} 
\end{figure*} 

\section{Introduction}

In the last decade, autonomous vehicles have gone from academic obscurity to one of the biggest industries in the technology sector. This is in no small part due to the rapid evolution of deep learning~\cite{paszkePyTorchImperativeStyle2019}, and improvements to autonomous vehicle (AV) simulation~\cite{dosovitskiy2017carla}. In fact in almost all modern systems for training or testing of autonomous vehicles, simulation plays a crucial role~\cite{schoner2018simulation}. This is because both training and testing in the real world is not only expensive and time-consuming, but represents a legitimate danger to public safety. Therefore academic and industry researchers have placed progressively growing effort into producing simulators which are more realistic and efficient to use. In many cases, these simulators have allowed both academic researchers as well as industry professionals to solve a number of difficult problems in autonomous control~\cite{mmdet3d2020}, and even computer vision~\cite{gaidon2016virtual}. However, even many modern simulators ignore the current technical requirements for both the training and testing of complex controllers, which may need to satisfy a range of different engineering requirements. More specifically, existing AV simulators often lack realistic traffic behaviour and are not easy to modify as requirements for training or testing change. Additionally, many of these simulators have ignored advances in computation graphs and deep-learning that make learning better policies easier~\cite{murthy2021gradsim}. 

To address these issues we first present \tds, a lightweight 2D driving simulator built entirely in PyTorch. \tds is differentiable and produces a computation graph which includes state transition functions based upon kinematic models and observation information produced via differentiable rendering. Additionally, the kinematic models which define the underlying transitions of \tds can be easily modified or customized, as can the rendered observations. Furthermore, metrics like collisions, off-road, and wrong way infractions are also fully differentiable. For ease of use in model training and evaluation, \tds provides first class support for batch processing, and is defined by a modular collection of wrappers which can change simulator behaviour. \tds also allows for more diverse traffic environments through the use of heterogeneous agent types such as vehicles, pedestrians, or cyclists which each have their own kinematic model. Access to control information such as traffic lights is also included in order to further improve realism.

Next we introduce the \tde benchmark suite, a training and testing environment for developing autonomous driving algorithms. Unlike many existing simulation environments, \tde provides access to realistic external agents via integration with a state of the art commercial (free and low cost academic licensing available) API, and provides a differentiable renderer through \tds. Additionally, \tde uses standard reinforcement learning (RL) environment construction through the use of step and reset functions following the Open-AI gym environment standard~\cite{brockmanOpenAIGym2016}, making integration with RL libraries (e.g.~stable baselines~\cite{stable-baselines3}) quick and easy. In contrast to most RL benchmark suites which train and test in the same environment, \tde also includes both validation and training environments in order to allow users to verify the generalization of their learned controllers. Lastly, \tde comes with access to its own environment labelling tool with which users can define their own unique AV scenarios to train and test models. This labelling tool allows the user to define information like static initial conditions or waypoint information, that can then be integrated with additional realistic, reactive, and diverse non-playable characters (NPCs) to create complex AV scenarios which test specific controller behaviors. 

In the following sections, we will illustrate how \tds functions, how it can be used or modified, and some of the advantages it offers over comparable simulators. We then introduce the \tde benchmark, and provide a set of standard RL baselines where we report both training and validation performance. We conclude with a discussion of future research directions.



 
\section{Related Work}
Recent years have seen a major surge in autonomous vehicle simulation, and this increased interest has led to progressively more realistic simulation environments~\cite{dosovitskiy2017carla}. In many cases these simulators were originally constructed to reduce the so called simulation to reality gap~\cite{tobin2017domain,osinski2020simulation}, which is characterized by how well controllers learned in simulation, function in a real world setting. Eventually, many of these simulators evolved into, or were adapted to machine learning and control benchmarks in order to test a number of different policy learning problems~\cite{zhan2019interaction}. Some simulators and benchmarks test performance in multi agent environments where external (non-ego) agents are executed using either rule based setups~\cite{xiao2021rule} or by replaying historical data~\cite{vinitsky2022nocturne}. Many of these multi-agent autonomous vehicle benchmarks have also been tailored for image based deep learning systems, and can test a variety of algorithm classes~\cite{muhammad2020deep}. In some cases, these benchmarks also include egocentric driving log data which can be used for learning via behavioural cloning~\cite{bansal2018chauffeurnet}. 
 
In other cases only local (non-expert) information is included, like the state-action reward~\cite{Sutton1998} or an infraction indicator which shows whether the ego vehicle experienced a collision, went off road, or is driving the wrong direction along a roadway. The algorithms which can be tested in this setting generally include some version of classical reinforcement learning~\cite{lu2023imitation}, or decision transformers~\cite{fu2022decision,dong2022development}. In some cases, when explicit waypoint information is available without direct expert supervision (e.g. the actions taken for each state), it is common to use either inverse reinforcement learning~\cite{igl2022symphony}, or to combine some form of closed loop control with a computer vision system~\cite{chen2020learning}. Such control systems also extend to trajectory prediction~\cite{nayakanti2023wayformer}, and given infraction information can be easily adapted to problems related to learning safe autonomous vehicle controllers~\cite{vitelli2022safetynet}. 

Structurally such benchmarks follow the example of \cite{brockmanOpenAIGym2016}, and include standardized step, reset, and initialization functions. This structure allows for ease and  accessibility, and has been adopted by the majority of more modern RL benchmarks~\cite{tunyasuvunakool2020}. Unfortunately, due to the fundamental difficulty surrounding RL (e.g. non-stationary, policy dependent data distributions~\cite{sutton2018reinforcement}), the majority of benchmarks test generalization. One example which does, is MetaWorld~\cite{yu2019meta}, a multi-agent robot manipulation environment. While MetaWorld allows the user to split learning into training and validation scenarios that test how well the learned policy generalizes to new objects and tasks, it is primarily tailored to Meta-RL~\cite{beck2023survey} and few-shot learning~\cite{10.1145/3582688}. Another popular example is ProcGen~\cite{cobbe2020leveraging}, a procedurally generated Atari-like gaming environment which generates levels which the agent must solve using random procedural generation. Both of these benchmarks present excellent test-beds for RL environments, however neither includes both data-driven scenario generation \textit{and} the train-validation scenario splits provided by \tde.

\begin{table*}[t]
\footnotesize
\setlength{\tabcolsep}{1mm}
\centering
\begin{tabular}{l|c|c|c|c|c|c|c|c}
\specialrule{.2em}{.1em}{.1em}
& Accel. & Sensor Sim & Expert Data & Sim-agents & Real Data & Routes/Goals & NPC type & License\\
\hline
SUMMIT~\cite{cai2020summit} & 
 & 
\cmark & 
 & 
\cmark &
\cmark &
- & Rule based & MIT\\
MACAD~\cite{palanisamy2020multi} & 
 & 
\cmark & 
 & 
\cmark &
 & 
Goal point & Jointly RL trained & MIT\\
DeepDrive-Zero~\cite{craig_quiter_2020_3871907} &
 & 
 &
 & 
\cmark &
 & 
- & Jointly RL trained  & MIT\\
SMARTS~\cite{pmlr-v155-zhou21a} & 
 & 
 & 
 & \cmark
 & 
 & 
Waypoints & Rule based & MIT\\
MADRaS~\cite{santara2021madras} & 
 & 
\cmark & 
 & 
\cmark &
 & 
Goal point & Rule based & GNU Affero\\
Nocturne~\cite{vinitsky2022nocturne}& 
 &
 &
\cmark & 
\cmark & 
\cmark & 
Goal point & Replay & MIT\\
MetaDrive~\cite{li2022metadrive} &
 &
\cmark &
\cmark &
\cmark &
\cmark &
- 
& Rule based + RL & Apache\\
Intersim~\cite{sun2022intersim} & 
 & 
 & 
\cmark & 
\cmark &
\cmark &
Goal point & Data driven (deterministic) &  MIT$^*$\\
\tds (Ours)& 
\cmark & 
 & 
 & 
\cmark &
 & 
Waypoints & Data driven (diverse) & Apache \\
tbsim~\cite{xu2022bits} & 
 & 
 & 
\cmark & 
\cmark &
\cmark & 
Goal point & Data driven (diverse) & Nvidia \\
\waymax~\cite{gulino2023waymax}  & 
\cmark & 
 &
\cmark & 
\cmark & 
\cmark &
Waypoints
& Replay + IDM & Waymax\\ 
\specialrule{.1em}{.05em}{.05em}

\end{tabular}
\caption{Reproduced in modified form with permission from the authors from \cite{gulino2023waymax}. Comparison of driving simulators in chronological order, with each column indicating the features available. \emph{Accel.}: In-graph compilation for hardware (GPU/TPU) acceleration. \emph{Sensor Sim}: Sensors (e.g. camera, lidar \& radar) input simulation.
\emph{Expert Data}: Human demonstrations or rollout trajectories collected with an expert policy.
\emph{Sim-agents}: agent models for simulated objects (e.g. other vehicles).
\emph{Real data}: Real world driving data.
\emph{Routes/Goals}: "$-$" means no routes or goals are provided; "Waypoints" means positions sampled from a trajectory; "Directions" means discrete driving directions including left, straight, and right; "Goal point" means the goal position. \emph{NPC type}: Reports the type of NPCs provided with the simulator. ``Data driven'' refers to modelled NPCs learned from data. \emph{License}: License type. MIT$^*$ indicates the simulator is MIT licensed, but the provided models are trained on Waymo licensed data. Unlike previous work we only include AV simulators which provide multi-agent simulated environments, for a more complete list see~\cite{gulino2023waymax}.}
\label{table:simulators} 
\end{table*}

\section{Design and Features}

In the following section we will describe the basic features available through \tds, then discuss the benchmark defined by \tde. Before getting into the specific details however, we briefly include a small code snippet which can be used to generate samples from the \tde environment:
\vspace{0.3cm}

\begin{adjustbox}{width=.45\textwidth,left}
\hspace{0.15cm}
\begin{python}  
# env = gym.make('tde/v0/validation') 
env = gym.make('tde/v0/train') 
observation, info = env.reset(seed=42)
for _ in range(1000):
    action = env.action_space.sample()  
    obs, reward, terminated, truncated, info = \
        env.step(action)
    if terminated or truncated:
      observation, info = env.reset() 
env.close()
\end{python}
\end{adjustbox}

\subsection{Simulator}
 
The \tds package can broken down into a number of different classes and helper functions, but the most important is the simulator class. This class defines a simulator object created using a static background, which by default includes both road and lane markings represented by a rendered triangle mesh. This simulator also includes control elements (e.g. traffic lights, yield signs) represented by rectangles that each have a local internal state whose semantics are not explicitly enforced. Each simulator object includes a collection of agents grouped by their type (car, pedestrian, or bicycle) which are represented with rigid rectangles. Each agent has a local kinematic model which defines its action space, and how actions from that agent will translate into motion. Lastly, each simulator contains a configurable render object, which can display the world from a birds eye view using a customizable colour map. For other examples of comparable AV simulators related to \tds see~\Cref{table:simulators}.

\subsection{Agents}

Each agent within the simulator is defined by a state. This state is described by set of static attributes (length, width, and anything else necessary to its kinematic models), a dynamic state (x, y, orientation, speed), and a flag (present mask) which indicates whether a given agent is currently active within the simulation. At each time step, all agents perform an action simultaneously which alongside the current agent state, determines their next state. \tds allows the agents to overlap, and defines this overlap as a collision event. Similarly \tds allows for overlap (or lack thereof) with any other bounding boxes or lane markings inside the mesh and identifies these events as other types of infractions (such as wrong-way or off-road).  

 
\begin{figure}[t!]
    \centering
    \vspace{0.245cm}
    \begin{subfigure}{0.235\textwidth}
       \includegraphics[width=\textwidth]{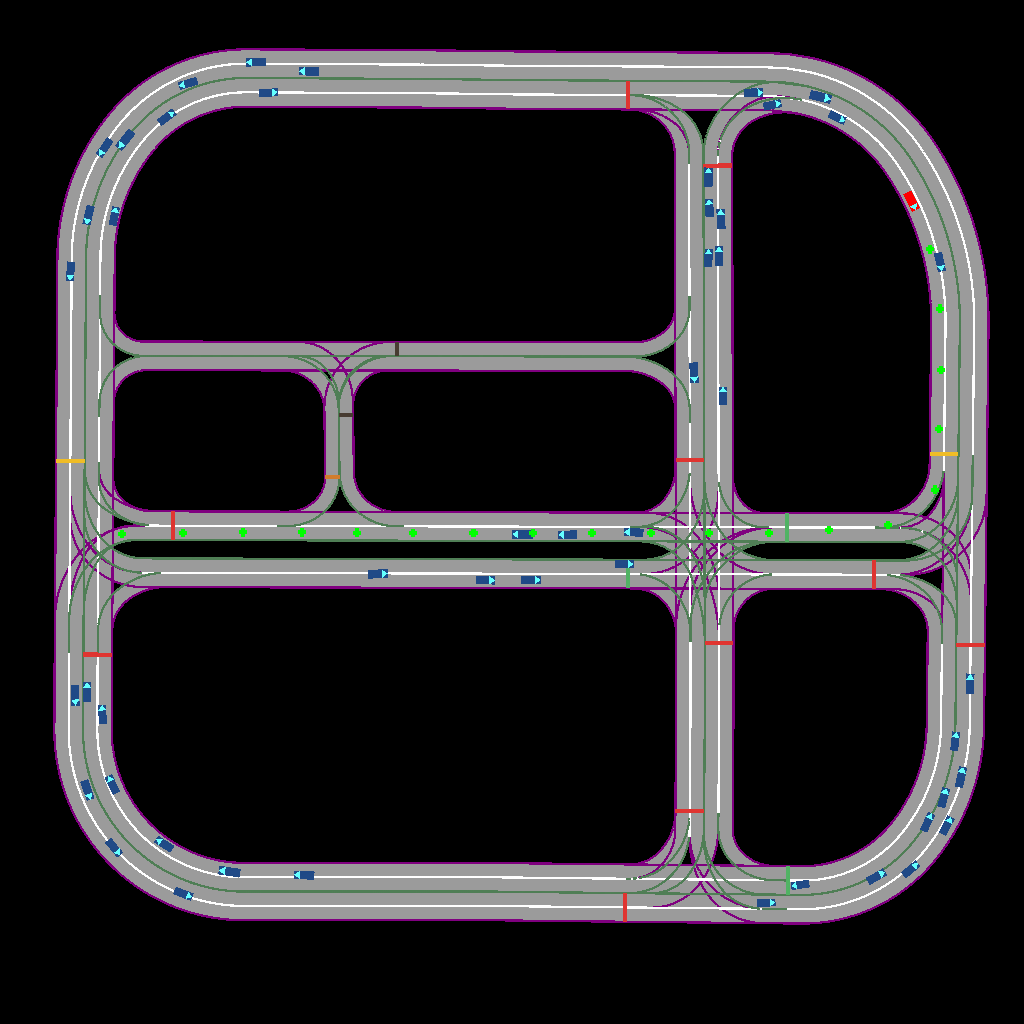}
       \caption{Scene 1, Initialization 1}
       \label{fig:train-ex:sc-1:init-1}
    \end{subfigure} \hfill
    \begin{subfigure}{0.235\textwidth}
       \includegraphics[width=\textwidth]{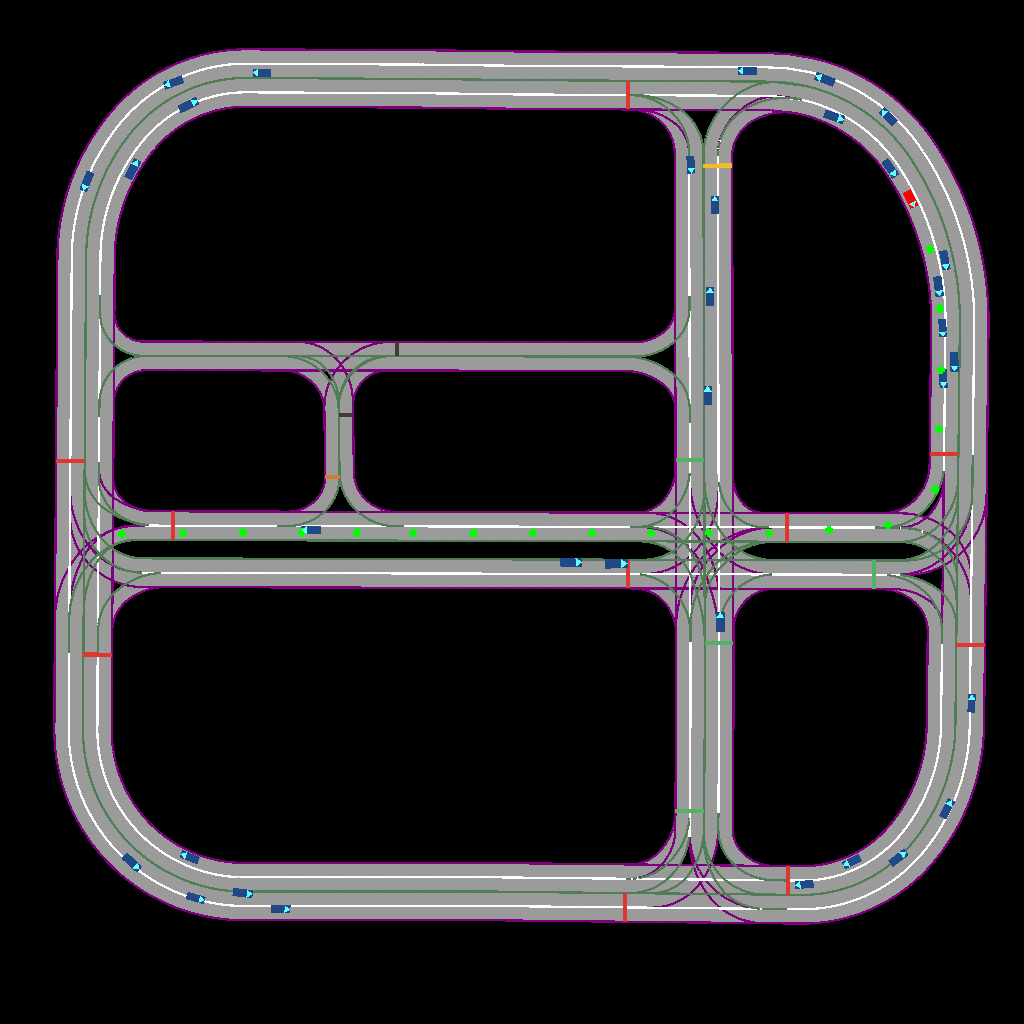} 
       \caption{Scene 1, Initialization 2}
       \label{fig:train-ex:sc-1:init-2}
    \end{subfigure} \\
     \vspace{0.2cm}
    \begin{subfigure}{0.235\textwidth}
       \includegraphics[width=\textwidth]{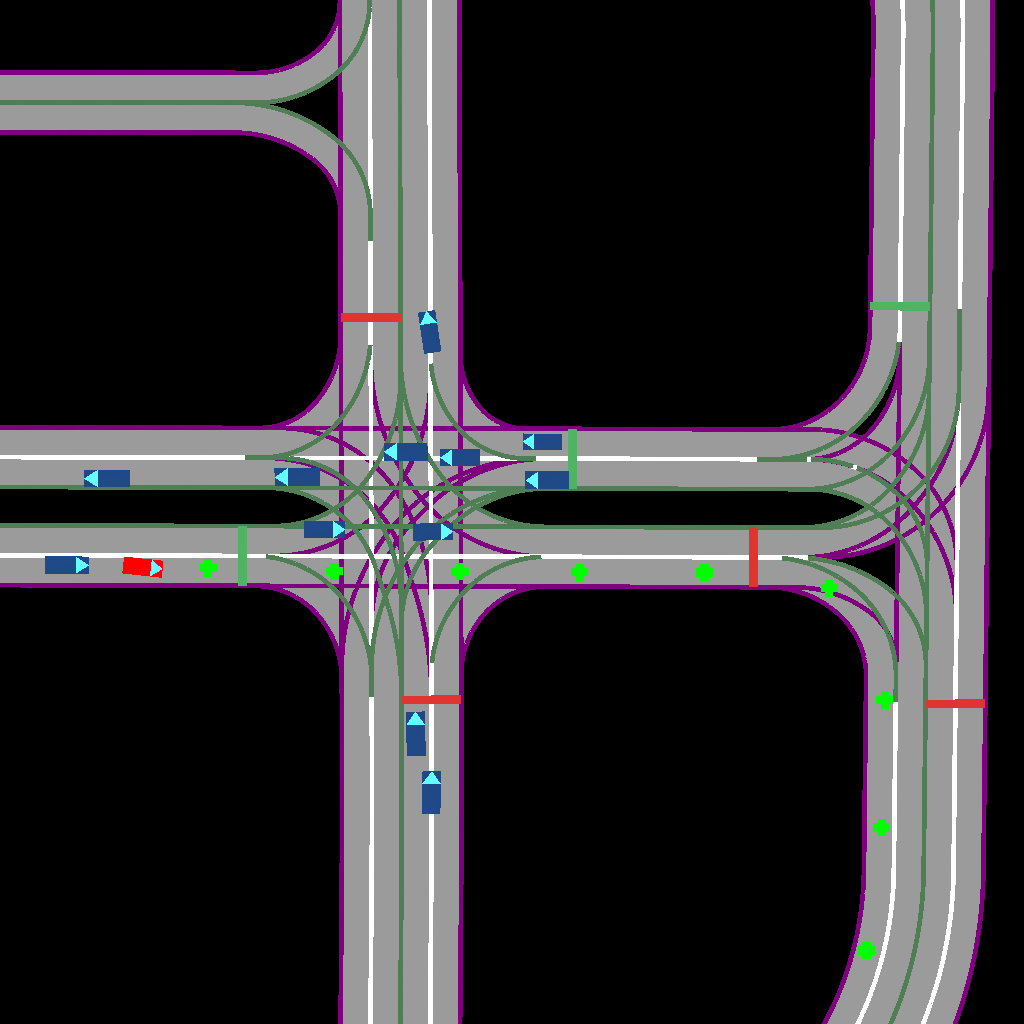}
       \caption{Scene 2, Initialization 1}
       \label{fig:train-ex:sc-2:init-1}
    \end{subfigure} \hfill
    \begin{subfigure}{0.235\textwidth}
       \includegraphics[width=\textwidth]{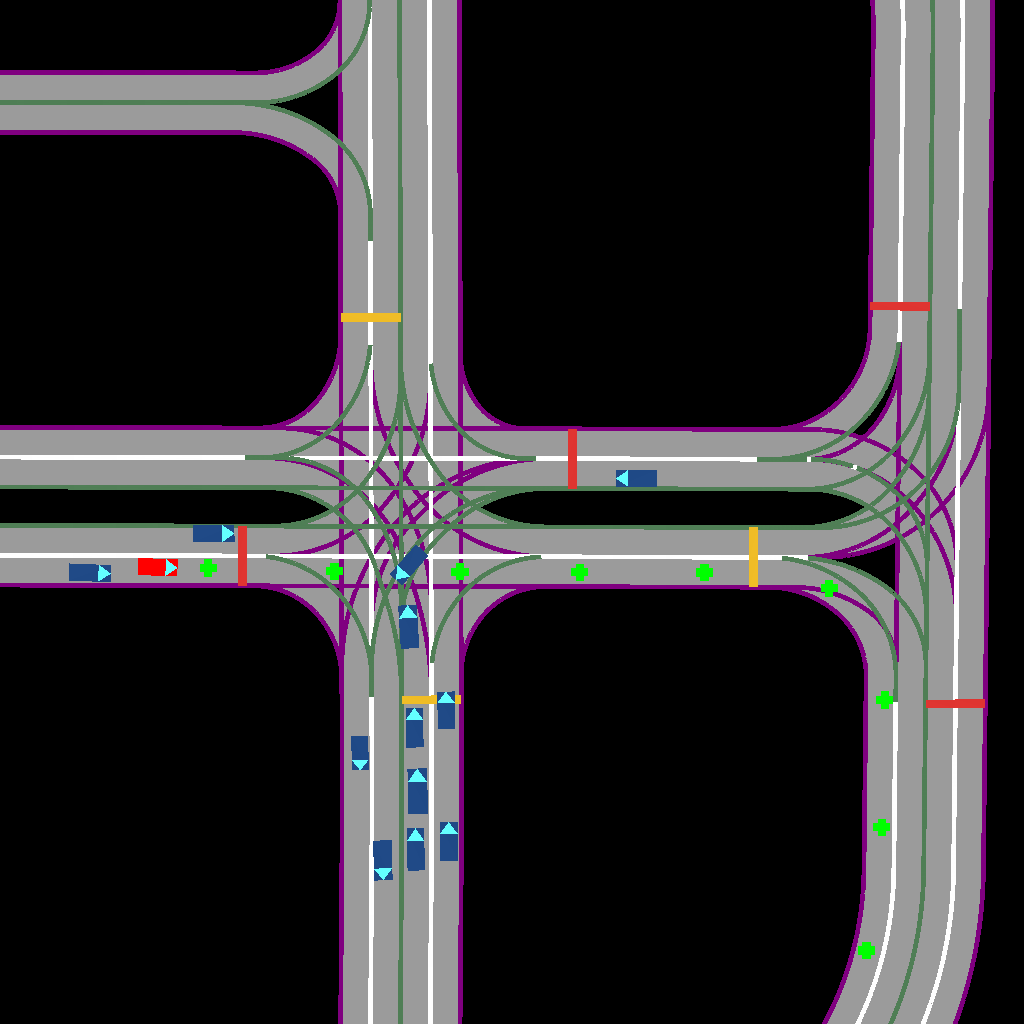} 
       \caption{Scene 2, Initialization 2}
       \label{fig:train-ex:sc-2:init-2}
    \end{subfigure} \\
    \caption{Two stochastic initializations drawn from two separate training scenarios. The first scene (\Cref{fig:train-ex:sc-1:init-1} and \Cref{fig:train-ex:sc-1:init-2}) includes the initialization over the entire map, while the second (\Cref{fig:train-ex:sc-2:init-1} and \Cref{fig:train-ex:sc-2:init-2}) provides a similar visualization of a specific intersection in a separate scenario. While certain distributional characteristics are similar (e.g. cars are stopped at a stop-lights), positions and even initial velocities of these vehicles differ between initializations. In all scenes, green dots indicate the sequence of waypoints provided to the agent, the red rectangle indicates the ego vehicle, blue rectangles indicate NPCs, and an orange arrow indicates the direction of the vehicles. Traffic lights are provided by coloured bars, which are red, yellow, or green. Notice that all NPCs in each of the initializations are going the correct direction and conform to traffic light state.}
    \vspace{-0.5cm}
    \label{fig:train-ex:inits}
\end{figure}
\tds and its extension \tde also include wrappers which can modify the simulator’s behaviour. Such wrappers can define which subset of agents will be controlled through replay or predefined controllers, or may remove agents that exit a designated area on the map. Wrappers for monitoring and visualization, which allow the user to log more common infractions like-off road or record videos of the scenes generated through agent interaction are also included. Unless otherwise specified, all wrappers are composable, and thus can be combined to achieve a wide range of user requirements. 

The default kinematic model used by all non-egocentric vehicles in \tds is the bicycle model~\cite{polackKinematicBicycleModel2017}, in which actions are defined by steering and acceleration. Other kinematic models are also available, including the unconstrained model where each action is defined by the change in state at subsequent time steps. We also include a teleporting variant where each action directly sets the next state. Tools and documentation are provided which illustrate how to modify these action models, or create completely custom, potentially more complex action models. The teleporting model for instance, is defined simply through inheritance by the following class:
\vspace{0.2cm}  

\begin{adjustbox}{width=.38\textwidth,left}
\hspace{0.115cm}
\begin{python}
class TeleportingModel(KinematicModel):
    def step(self, action, dt=None):
        self.set_state(action)
    def fit_action(self, future_state, \
            current_state=None, dt=None):
        return future_state
\end{python}
\end{adjustbox}

\vspace{0.2cm} 
In this example, the state is set directly through the action passed to the kinematic models step function. While this kinematic model is simple, extending this framework more complex kinematic models follows in a similar fashion.

\subsection{Generative Models of Vehicle Behaviour}

Provided with a known kinematic model, avoiding basic infractions like off-road or wrong-way behaviour can often be accomplished through methods in classic control~\cite{lavalle2006planning}. The more difficult scenarios which most practitioners and researchers are interested in however, are those which contain non-static state information like changing traffic lights, or include external agents (NPCs). While environment attributes like traffic lights can be controlled realistically through simple state machines, inclusion of NPCs within an environment requires \emph{realistic}, \emph{reactive}, and \emph{diverse} models of driving behaviour~\cite{scibiorImaginingRoadAhead2021}. Through integration with with an API that serves agent initializations and behaviour predictions, we extend \tds and introduce \tde. Through \tde, complex behavioural models can be directly queried to provide a service that initializes and controls NPCs. Through this integration, expensive AV models and (the GPUs required to run them) can be abstracted behind a simple API layer, thereby reducing the need for increased local computing resources.
  
The two primary API endpoints which \tde uses are \initlink and \drivelink. As is illustrated in \Cref{fig:train-ex:inits}, \initlink is called during simulator initialization to populate the scenario with a randomized but realistic collection of NPCs, and is based upon recent work in conditional diffusion generative modeling~\cite{zwartsenberg2023conditional, naderiparizi2023dont}. Notably, the user can also choose to manually set the initializations of all, or a subset of NPCs, through modifications of the environment's associated definition. During initialization, any user-defined agents are placed in the scene first, after which, additional agents are sampled conditionally using locations of the manually placed agents. Conditionally sampled agents can have their agent type specified, but are otherwise set as basic vehicles. We note that this conditional (or unconditional) initialization of external agents is traffic-light state aware. After all agents have been initialized, \drivelink is then called during the environments step function, and provides driving behaviors for each of the initialized NPCs present within the scene using recent advances in models of human realistic vehicle behaviors~\cite{scibiorImaginingRoadAhead2021}.

\begin{figure*}[h!]
    \centering 
    \includegraphics[width=0.8\textwidth]{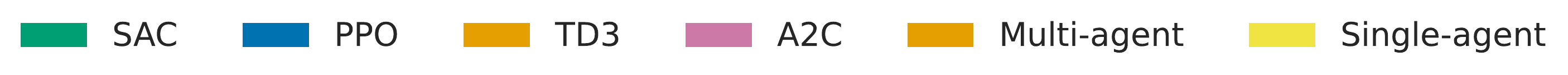}
    \begin{subfigure}[b]{1.0\textwidth}
         \centering \includegraphics[width=0.995\textwidth]{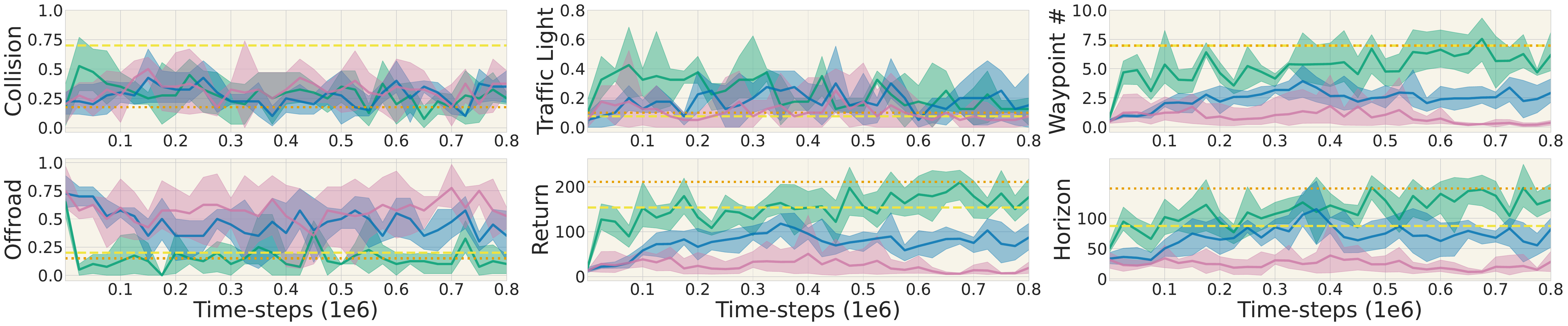}  
         \vspace{-0.5cm}
         \caption{Multi-Agent-Train}
         \label{fig:single-agent:train}
     \end{subfigure}
    \begin{subfigure}[b]{1.0\textwidth}
         \centering \includegraphics[width=0.995\textwidth]{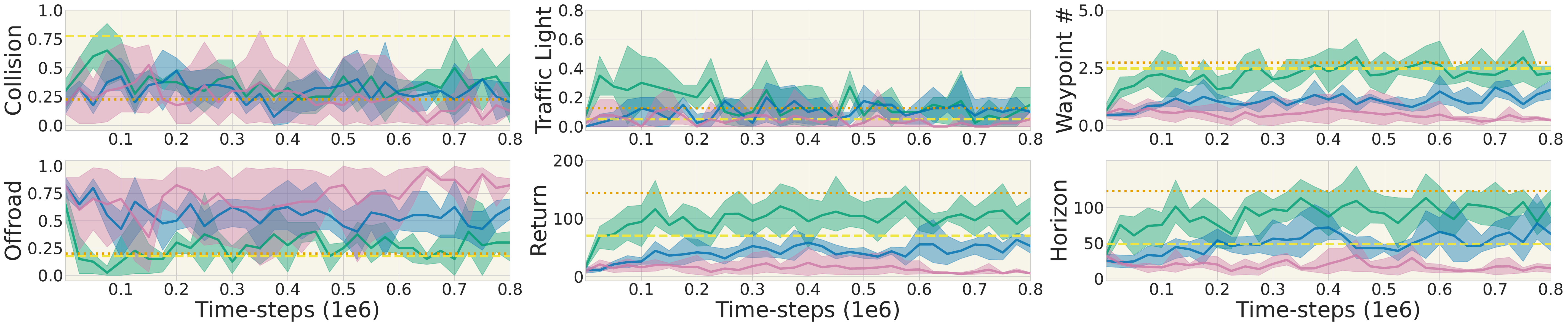} 
         \vspace{-0.5cm}
         \caption{Multi-Agent-Validation}
         \label{fig:single-agent:val}
     \end{subfigure}  
     \begin{subfigure}[b]{1.0\textwidth}
         \centering \includegraphics[width=0.995\textwidth]{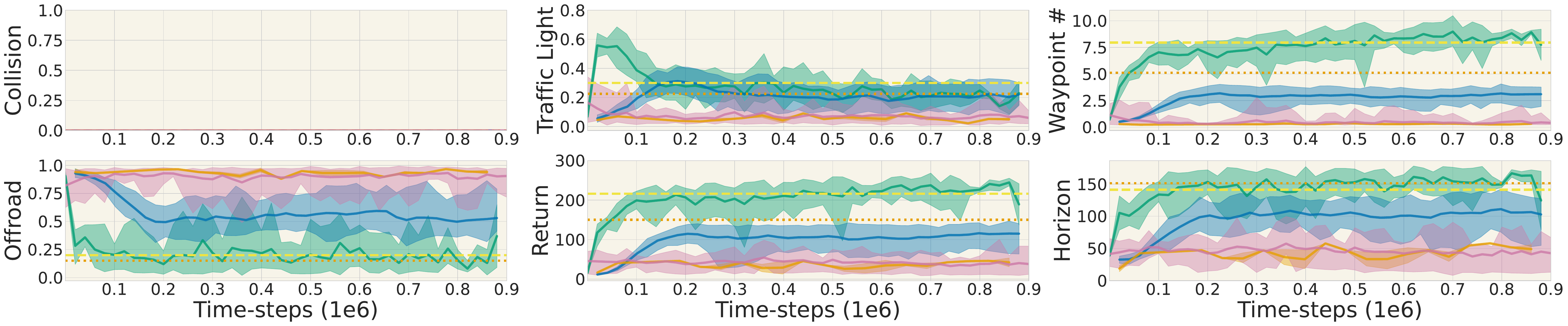}
         \vspace{-0.5cm}
         \caption{Single-Agent-Train}
         \label{fig:multi-agent:train}
     \end{subfigure}
    \begin{subfigure}[b]{1.0\textwidth}
         \centering \includegraphics[width=0.995\textwidth]{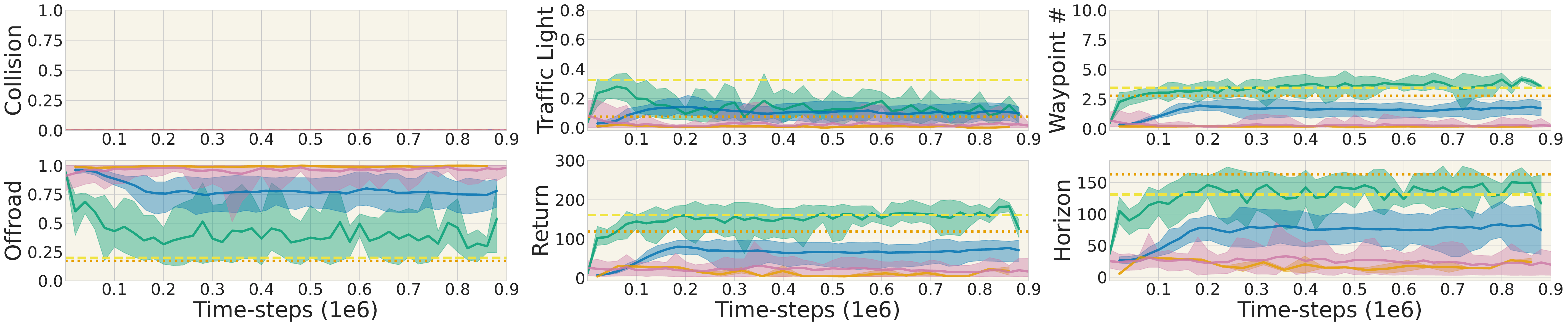}
         \vspace{-0.5cm}
         \caption{Single-Agent-Validation}
         \label{fig:multi-agent:val}
     \end{subfigure}
    \caption{The \tde benchmark: training curves for multi-agent and single-agent \tde environments across each of their respective training and validation scenarios. \emph{Collision} indicates the percentage of time in which an episode ends in a collision, \emph{Traffic-Light} indicates the percentage of times an episode ends in a traffic light violation, \emph{Waypoint \#} indicates the average number of waypoints achieved during an episode, \emph{Offroad} indicates the percentage of episodes which end in an off road infraction, \emph{Return} indicates the average cumulative reward per episode,  and finally \emph{Horizon} indicates the average length of episodes observed. We include two dotted and dashed lines which indicate the average performance of across 4 randomly seeded RL agents evaluated on 10 episodes each. The dotted line indicates performance of agents trained in a multi-agent environment, while the dashed line indicates performance of agents  trained in an ego-only environment. The plots above illustrate two important things. The first is that learning agents within a multi-agent environment is more difficult then in a single agent environment, as illustrated by significantly lower reward and time-horizon metrics. Second the performance gap between multi-agent trained models in a single agent environment is generally much lower (see the horizon and return plots) then single-agent trained models being evaluated in a multi-agent setting. Across all benchmarks, vehicles trained in a multi-agent environment survive for longer (larger horizon) than ego only trained environments.}
    \vspace{-0.2cm}
    \label{fig:train-plots}
\end{figure*}

 
\subsection{Environment}
As described above, the \tde package bundles \tds with API based behavioural models for initialization and reactive NPCs. Using this environment as its basis, \tde then defines scenarios with which various RL and AV algorithms can be trained or tested. For this initial \tde benchmark release all maps come from the CARLA simulator~\cite{dosovitskiy2017carla}. Each of these training and testing environments are set up such that the user controls only the actions of the ego agent within a given scenario, while \drivelink controls the remaining agents. Similarly, while additional NPCs are generated through \initlink, the ego agent is randomly  initialized locally along a scenario-specific sequence of ego-initialization waypoints. \tde is also augmented with a reward based on infraction metrics and waypoints, comes with pre-defined training and validation scenarios, and has the option to use external data and a scenario editor to create hand tailored new scenarios for whatever the user may wish to test. The core \tde environment is based on OpenAI Gym~\cite{brockmanOpenAIGym2016}, and thus can be easily modified through wrappers which follow standard function. For example, one could easily modify the state observations or reward by simply creating their own environment wrapper:
\vspace{0.3cm} 

\begin{adjustbox}{width=.45\textwidth,center}
\begin{python}
class MyTorchDriveEnv(gym.Wrapper): 
    def __init__(self, env: gym.Env):
        super().__init__(env)
    def reset(self, **kwargs):
        obs, _ = super().reset(**kwargs) 
        return self.transform_out(obs), _
    def step(self, action: Tensor): 
        obs, reward, terminated, trunc, info \ 
        = super().step(action)
        return obs, reward, terminated, trunc, info
    def render(self, *args, **kwargs):
        return self.env.render(*args, **kwargs)
    def close(self):
        self.env.close()
\end{python} 
\end{adjustbox}
\vspace{0.3cm}

For the set of standard training and validation benchmarks, we break down each of the attributes which define the reinforcement learning problem below.

\subsubsection{Action Space}
The action space defines the set of admissible actions which can be taken by a controller. Within \tde these actions are continuous, and defined by a steering angle $b$, and acceleration $c$ such that $b \in (-0.3, 0.3) $ and $c \in (-1.0, 1.0)$. Constraints for $b$ are chosen to ensure vehicle behaviour remains visually realistic. 

\subsubsection{Observation Space} The observation space within the \tde environment is given by a 2d rendering of the egocentric birds-eye view of the scene (see \Cref{fig:train-ex}), i.e.~the ego agent is always facing the same direction and the map translates and rotates. For each benchmark, this rendered observation is provided using a tensor of size $3 \times 3 \times 64 \times 64$, where the first index of tensor indicates historical frames from the previous two time-steps plus the current observation. 

\subsubsection{End-Conditions} Like most popular RL environments \tde defines a set of conditions which if met, will halt simulation. These end conditions include: if the ego agent goes off-road, if there is a traffic light violation (the ego agent passes over a traffic-light bounding box), and finally if the ego agent collides with another agent within the scene. We also include a maximum timesteps end condition to ensure metrics and evaluation can be computed efficiently. 


\subsubsection{Reward} In this setting, because we implicitly limit the maximum cumulative return through the end conditions defined above, we include only simple feedback via the reward which penalizes based upon a notion of smoothness, and rewards based upon how accurately the agent navigates between waypoints. Within \tde, these waypoints define the general path which the agent should take, and are denoted by a small circle rendered on top of the rendered birdview observations. In order to reduce the need for strong exploration, we also include a small bonus for when the agent moves some minimum distance between different timesteps. Within the benchmark suite, the reward is defined using two bonuses (strictly non-negative values), and a penalty (a strictly non-positive value). This gives a simple reward $r$ defined using three values, 
\begin{align*}
    r &= \alpha_1 |\text{Movement Bonus}| + \alpha_2 |\text{Waypoint Bonus}| \\
    & - \beta_1 |\text{Smoothness Penalty}|, 
\end{align*}
where for the benchmark results shown in~\Cref{fig:train-plots}: $\alpha_1=1$, $\alpha_2=10$, and $\beta_1=0.05$. CSV files containing the data used to generate the plots can be found on the accompanying github repository. We also note that a movement bonus is only included if the agent travels at least 0.5 meters within a given time-step. Similarly, the waypoint bonus is only included if the agent transitions to a position which is within two meters of the next waypoint in the queue. At which point the waypoint is popped from the queue and the next one is placed in the environment as a new green dot. The smoothness penalty is defined by the cosine similarity between headings at the current time-step and the next time-step.

\section{Benchmark}

In order to illustrate the use of \tde as a reinforcement learning (RL) benchmark, we have fully integrated the stable-baselines~\cite{stable-baselines3} library in order evaluate the performance a number of different policy learning algorithms. These algorithms include A2C~\cite{mnih2016asynchronous}, PPO~\cite{schulman2017proximal}, SAC~\cite{haarnoja2018soft}, and TD3~\cite{pmlr-v80-fujimoto18a}. We report a number of different metrics available to the user through the \tde interface alongside the base return (cumulative reward), including infraction information, the average time-horizon observed by the agent, and the average number of waypoints achieved. \Cref{fig:single-agent:train} represents a training environment where there are no external agents during training, while \Cref{fig:multi-agent:train} represents the training environment which contains reactive external agents. As expected the inclusion of additional agents dramatically increases the difficulty of the environment. Unlike many standard RL benchmarks, we also include a validation environment for both the single (\Cref{fig:single-agent:val}) and multi-agent (\Cref{fig:multi-agent:val}) settings. Notably, the two most performant algorithms are SAC and PPO across all benchmarks. To illustrate the importance of including multi-agent simulation in AV training, we also include dotted and dashed lines which define the average performance across models trained using the best performing (in terms of reward) RL algorithm within single agent and multi-agent environments across both training and testing environments. As expected, agents trained within the multi-agent environment tend to function reasonably well in both single agent and multi-agent environments, while agents trained using only single agent simulation do not generalize to the multi-agent setting.  


\section{Discussion}
In this paper we introduce \tds and its benchmark library \tde. We illustrate ease of use through both visual and program examples, and provide results on the \tde benchmark using \cite{stable-baselines3}. This benchmark confirms the importance of reactive and realistic NPCs in the training of AV controllers by considering the effects of out of distribution error through a validation dataset, as well as through the comparison of agents trained using only single agent interactions. We hope that through the introduction of this benchmark, researchers will work towards developing controllers that are more robust changes in distribution. Additionally, the results using standard RL baselines on this benchmark illustrate the need for a more precise treatment of performance than is provided by the average return. This is because, regardless of how high the reward in examples from \tde became for standard baselines, even the highest performing algorithms maintained non-trivial collision and off-road infractions. Actively designing algorithms that target both reward maximization while driving infractions to zero represents an interesting area of future research. In the future, we also wish to expand the baselines we consider to include: expert trained imitation learning models, RL algorithms with pretrained feature encoders, and finally policy learning algorithms which can take advantage of \tds's differentiable dynamics.




 \bibliography{main}
\bibliographystyle{unsrt} 

\end{document}